\documentclass{article}
\usepackage{placeins}
\usepackage{graphicx}   
\usepackage[utf8]{inputenc} % allow utf-8 input
\usepackage[T1]{fontenc}    % use 8-bit T1 fonts
\usepackage{booktabs}       % professional-quality tables
\usepackage{microtype}      % microtypography

\usepackage{amsfonts}       % blackboard math symbols
\usepackage{amssymb}        % math symbols
\usepackage{amsmath}        % split equations
\usepackage{nicefrac}       % compact symbols for 1/2, etc.
\usepackage{bm}             % bold math symbols
\usepackage{comment}        % For comments
\usepackage{soul}           % strikethrough text

\usepackage{enumitem}       % Removing indentation of enumerate
\usepackage{todonotes}
\makeatletter
\let\NAT@parse\undefined
\makeatother
\usepackage{verbatim}

\usepackage{hyperref}                  

\usepackage{todonotes}
\usepackage[accepted]{icml2019}

\icmltitlerunning{Super-resolution of Time-series Labels for Event Detection}

\begin{document}

\twocolumn[
\icmltitle{Super-resolution of Time-series Labels for Bootstrapped Event Detection}



\icmlsetsymbol{equal}{*}

\begin{icmlauthorlist}
\icmlauthor{Ivan Kiskin}{equal,ox}
\icmlauthor{Udeepa Meepegama}{equal,oxphys}
\icmlauthor{Stephen Roberts}{ox}
\end{icmlauthorlist}

\icmlaffiliation{ox}{Machine Learning Research Group, University of Oxford, Oxford, United Kingdom}
\icmlaffiliation{oxphys}{Department of Physics, University of Oxford, Oxford, United Kingdom}

\icmlcorrespondingauthor{Ivan Kiskin}{ikiskin@robots.ox.ac.uk}

\icmlkeywords{Machine Learning, ICML}

\vskip 0.3in
]



\printAffiliationsAndNotice{\icmlEqualContribution} % otherwise use the standard text.

\begin{abstract}
% Be concise
Solving real-world  problems, particularly with deep learning, relies on the availability of abundant, quality data. In this paper we develop a novel framework that maximises the utility of time-series datasets that contain only small quantities of expertly-labelled data, larger quantities of weakly (or coarsely) labelled data and a large volume of unlabelled data. This represents scenarios commonly encountered in the real world, such as in crowd-sourcing applications. In our work, we use a nested loop using a Kernel Density Estimator (KDE) to super-resolve the abundant low-quality data labels, thereby enabling effective training of a Convolutional Neural Network (CNN). We demonstrate two key results: a) The KDE is able to super-resolve labels more accurately, and with better calibrated probabilities, than well-established classifiers acting as baselines; b) Our CNN, trained on super-resolved labels from the KDE, achieves an improvement in F1 score of 22.1\,\% over the next best baseline system in our candidate problem domain.
\end{abstract}

\section{Introduction}
Finely-labelled data are crucial to the success of supervised and semi-supervised approaches to classification algorithms. In particular, common deep learning approaches \cite{NueralNet, NN} typically require a great number of data samples to train effectively \cite{DNNSmallData1, DNNSmallData2}.  In this work, a \emph{sample} refers to a section taken from a larger time-series dataset of audio. Often, these datasets lack large quantities of fine labels as producing them is extremely costly (requiring exact marking of start and stop times). The common distribution of data in these domains is such that there are small  quantities  of  expertly-labelled (finely)  data,  large  quantities  of  \textit{weakly} (coarsely)  labelled  data, and a large volume of unlabelled data. Here, \textit{weak} labels refer to labels that indicate one or more events are present in the sample, although do not contain the information as to the event frequency nor the exact location of occurrence(s) (illustrated in Section \ref{FeatureExtract}, Fig. \ref{fig:LogMel}). Our goal therefore is to  improve  classification  performance in domains with variable quality datasets. 

% In this paper we address time-series in the form of audio data. The two most common approaches to this domain are HMMs \cite{HMM} and  deep learning methods [cite]. With the rapid development in computational power and software, data-hungry deep learning methods have become the preferred approach \cite{SuperCNN1,SuperCNN2}. 
%However, as collecting these fine labels is very time consuming, the training data size is often limited to a few minutes or hours \cite{TUT, TUT1}. On the other hand, weakly labelled data takes much less time to annotate manually, since the annotator has to mark only the active sound event classes and not the exact event time boundaries.

% frames in a recording are regarded as instances in a bag, and the presence or absence of events is only known on the bag level.

%In \citet{WeakLabel1}, to approximate the MIL framework, a time-distributed CNN with a global max-pooling layer is employed to predict the temporal locations of each event. Whilst, in \cite{WeakLabel2}, a stacked convolutional and recurrent neural network (CRNN) architecture with two prediction layers is proposed for the same task. These weakly supervised methods suffer from inherent noise due to the weak labels.

Our key contribution is as follows. We propose a framework that combines the  strengths of both traditional algorithms and deep learning methods, to perform multi-resolution Bayesian bootstrapping. We obtain probabilistic labels for pseudo-fine labels, generated from weak labels, which can then be used to train a neural network. For the label refinement from weak to fine we use a Kernel Density Estimator (KDE).

The remainder of the paper is organised as follows. Section \ref{Methodology} discusses the structure of the framework as well as the baseline classifiers we test against. Section \ref{Experiments} describes the datasets we use and the details of the experiments carried out. Section \ref{Classification Performance} presents the experimental results, while Section \ref{Conclusion} concludes.

\section{Methodology}
\label{Methodology}

\subsection{Framework Overview}
\begin{figure}[]
\begin{center}
\includegraphics[width=7cm]{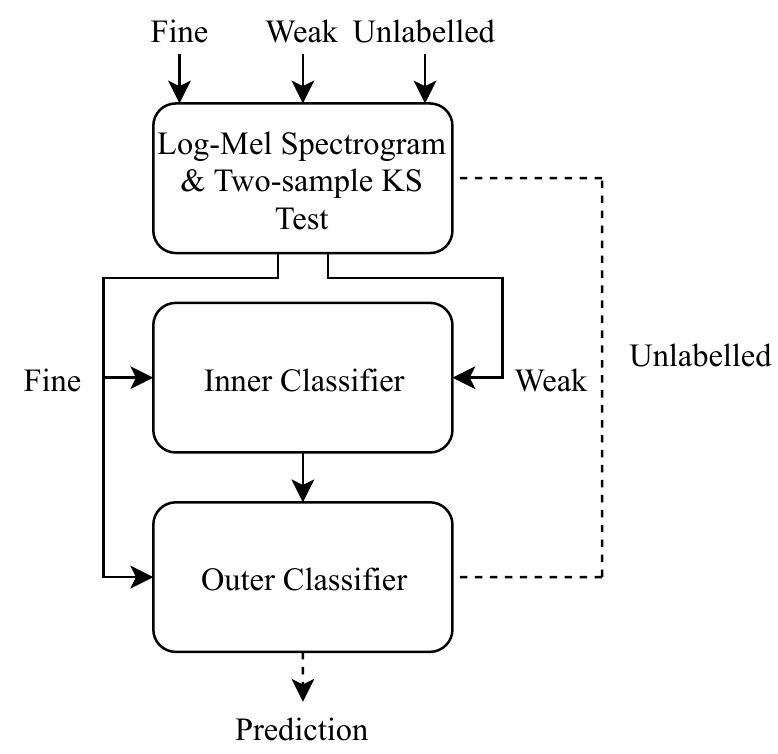}
\vspace{-0.1in}
\caption{Framework comprising a feature extraction \& selection layer, an inner classifier and an outer classifier. The arrows represent data flows.} 
\label{fig:Framework}
\end{center}
\end{figure}

Our framework is separated into an inner and outer classifier in cascade as in Figure \ref{fig:Framework}. For the inner classifier we extract features from the finely and weakly-labelled audio data using the two-sample Kolgomogrov-Smirnov test for features of a log-mel spectrogram (Section \ref{FeatureExtract}). We train our inner classifier, the Gaussian KDE (Section \ref{GaussianKDE}), on the finely-labelled data and predict on the weakly-labelled data. 

For the outer classifier we extract the feature vectors from an unlabelled audio dataset using the log-mel spectrogram. We then train our outer classifier, a CNN, (Section \ref{CNN}) on the finely-labelled data and the resulting pseudo-finely labelled data output by the Gaussian KDE. The details of our data and problem are found in Section \ref{Experiments}. Code will be made available on  \url{https://github.com/HumBug-Mosquito/weak_labelling}.

\subsection{Feature Extraction and Selection}
\label{FeatureExtract}
\begin{figure}
\begin{center}
\includegraphics[width=8.2cm]{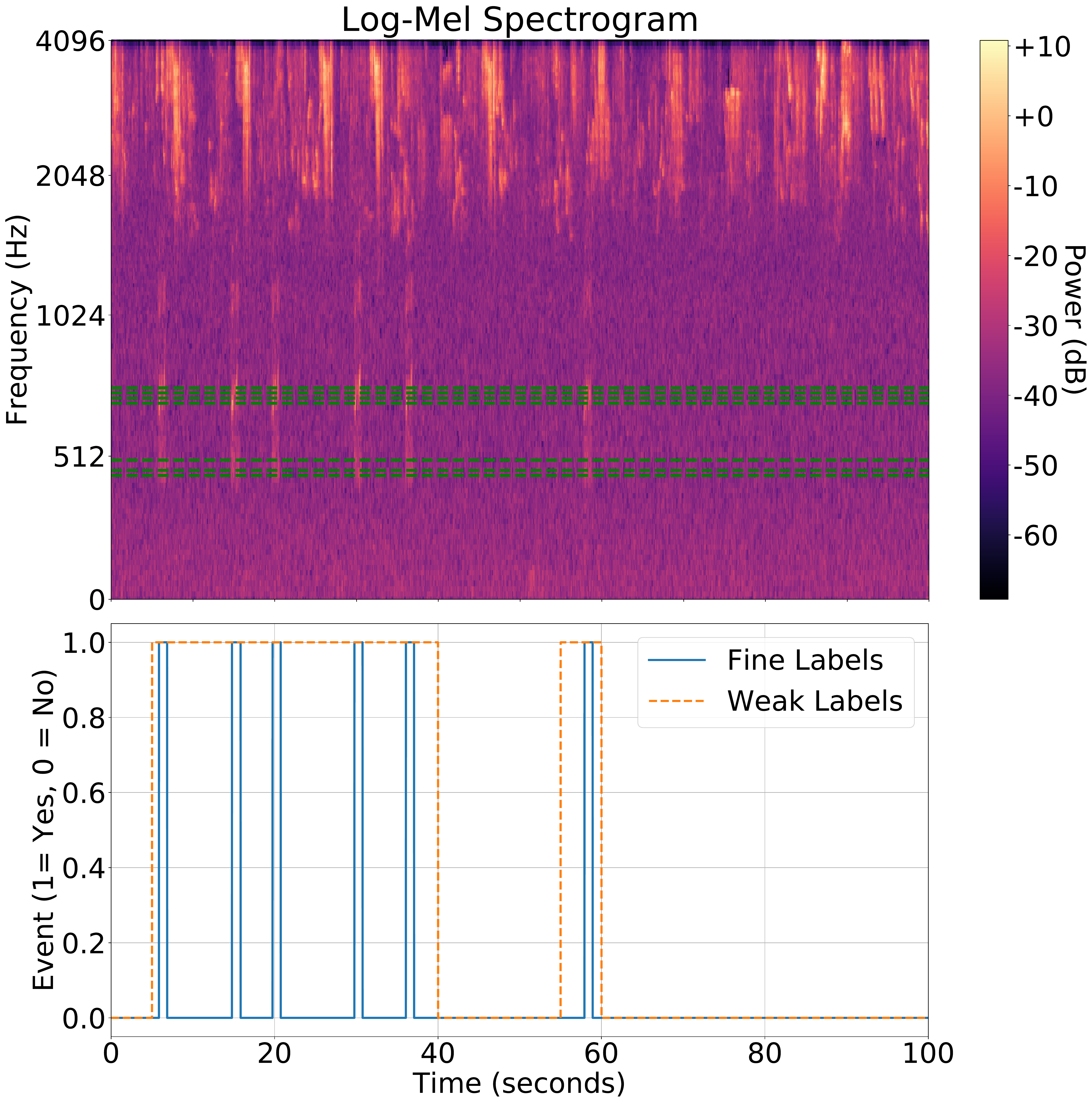}
\vspace{-0.1in}
\caption{From top to bottom: Log-mel spectrogram of $100$ seconds of audio data at a signal-to-noise ratio of $-15$ dB. The KS-selected features are shown as green dashed lines; The corresponding fine and weak labels for the above log-mel spectrogram.}
\label{fig:LogMel}
\end{center}
\end{figure}

The CNN uses the log-mel spectrogram (as in Fig.~\ref{fig:LogMel}) as it has recently become the gold standard in feature representation for audio data \cite{LogMel,LogMel4}.
%Previous state-of-the-art feature extraction methods such as Mel-frequency cepstral coefficients (MFCCs) \cite{MFCC} or Perceptual Linear Prediction (PLP) coefficients \cite{PLP} were largely hand-crafted for use with linear models. However, such features have become less popular with deep learning algorithms as it is no longer essential for feature representations to be sufficiently de-correlated \cite{LogMel1}. Additionally, the strengths of CNNs lie with their ability to learn localised patterns through weight-sharing and pooling operations \cite{LogMelCNN} -- patterns present in the features of spectrograms. 
The input signal is divided into $0.1$ second windows and we compute $128$ log-mel filterbank features. Thus, for a given $100$ seconds of audio input, the feature extraction method produces a $1000\times128$ output.

The two-sample Kolmogorov-Smirnov (KS) test \cite{KS1} is a non-parametric test for the equality of continuous, one-dimensional probability distributions that can be used to compare two samples. This measure of similarity is provided by the Kolmogorov-Smirnov statistic which quantifies a distance between the empirical distribution functions of the two samples. We use the KS test to select a subset of the $128$ log-mel features, that are maximally different between the two classes to feed into the classifiers. We choose $N$ features with the largest KS statistics. Fig. \ref{fig:LogMel} illustrates that the process to find maximally different feature pairs, correctly chooses frequencies of interest.  For example, if the noise file is concentrated in high frequencies (as in Fig. \ref{fig:LogMel}), the KS feature selection process chooses lower harmonics of the training signal (a mosquito flying tone)  as  features  to feed  to  the  algorithms.   Conversely, for  low-frequency  dominated  noise,  higher  audible  harmonics  of  the event signal are identified.

\subsection{Gaussian Kernel Density Estimation}
\label{GaussianKDE}

Kernel density estimation (KDE) or Parzen estimation \cite{KDE1, KDE2} is a non-parametric method for estimating a $d$-dimensional probability density function $f_{\mathbf{X}}(\bm{x})$ from a finite sample $\mathcal{D}=\{\bm{x}_{i}\}^{N}_{i=1}$, $\bm{x}_{i}\in\mathbb{R}^{d}$, by convolving the empirical density function with a kernel function.

We then use Bayes' theorem to calculate the posterior over class $1$
\begin{equation}\label{eq:Bayes}
    p(C_{1}\mid\bm{x})=\frac{p(\bm{x}\mid C_{1})p(C_{1})}{p(\bm{x}\mid C_{0})p(C_{0})+p(\bm{x}\mid C_{1})p(C_{1})},
\end{equation}
where $ p(\bm{x}\mid C_{k})$ is the KDE per class $C_k$, with $C_1$ representing the event class and $C_0$ the noise class (i.e. non-event). 

\subsection{Convolutional Neural Network}
\label{CNN}
\begin{figure}[]
\begin{center}
\includegraphics[width=8.2cm]{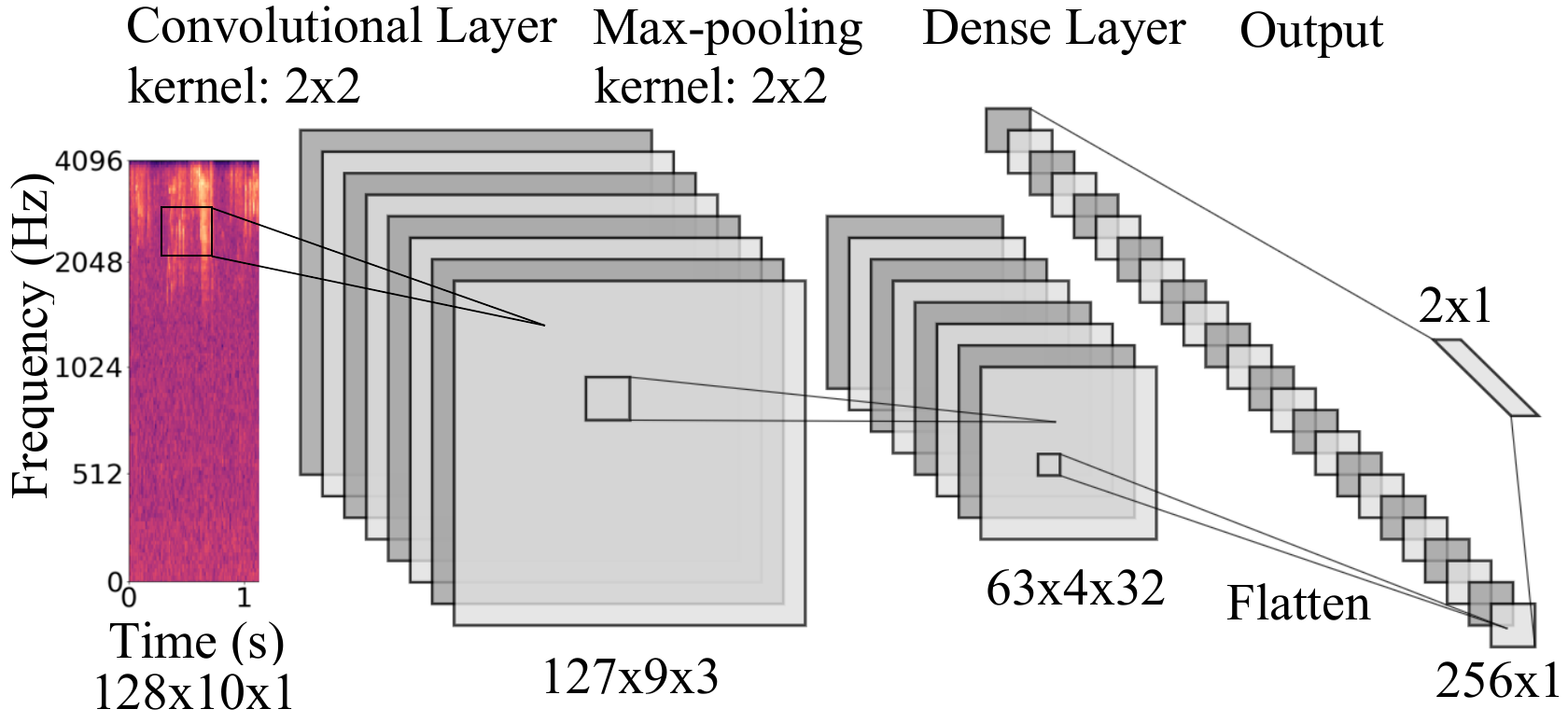}
\vspace{-0.1in}
\caption{The CNN architecture. Spectrogram of mosquito recording fed as input to convolutional layer with $32$ filters and kernel $2\times 2 \times 1$. Generated feature maps are down-sampled by a max-pooling layer from $127\times 9$ to $63\times 4$. It is then connected to a dense layer with $256$ neurons and finally the output with $2$ neurons.}
\label{fig:CNN}
\end{center}
\end{figure}

With our scarce data environment we use a CNN and dropout with probability $p=0.2$ \cite{Dropout}. Our proposed architecture, given in Fig. \ref{fig:CNN}, consists of an input layer connected sequentially to a single convolutional layer and a fully connected layer. The CNN is trained for $10$ epochs with SGD \cite{SGD}, and all activations are ReLUs. We use this particular architecture due to constraints in data size \cite{Wavelet} and therefore have few layers and fewer parameters to learn. 

\subsection{Traditional Classifier Baselines}
\label{BaseLine}
We compare our inner classifier, the Gaussian KDE, with
more traditional classifiers that are widely used in machine learning: Linear Discriminant Analysis (LDA), Gaussian Na\"ive Bayes (GNB), support vector machines using a radial basis function kernel (RBF-SVM), random forests (RF) and a multilayer perceptron (MLP).

\section{Experiments}
\label{Experiments}

\subsection{Description of Datasets}
\label{Data}

% Talk about crowd-sourcing
The most common scenario where mixed quality labels can be found is in crowd-sourcing tasks \cite{CrowdSourcing1, CrowdSourcing2, CrowdSourcing3}, or any challenge where data collection is expensive. The HumBug \cite{Zooniverse}, \cite{HumBug} project utilises crowd-sourcing, forming the main motivation for this research, as well as the basis for our signal\footnote{The overall goal of HumBug is real-time mosquito species recognition to identify potential malaria vectors in order to deliver intervention solutions effectively.}. The event signal consists of a stationary real mosquito audio recording with a duration of $1$ second. The noise file is a non-stationary section of ambient forest sound. The event signal is placed randomly throughout the  noise  file  at  varying  signal-to-noise ratios (SNRs), to create quantifiable data for prototyping  algorithms. There is a class imbalance of $1$ second of event to $9$ seconds of noise in the finely-labelled data and this is propagated to the weakly-labelled and unlabelled datasets. We include $100$ seconds of expert, finely-labelled data, $1000$ seconds of weakly-labelled data, and a further $1000$ seconds of unlabelled test data. To report performance metrics, we create synthetic labels at a resolution of $0.1$ seconds for the finely-labelled data, and on the order of $5$ seconds for the weakly-labelled data. We choose $5$ seconds as to allow the labeller to have temporal context when classifying audio as an event or non-event. As the listener is presented randomly sampled (or actively sampled \cite{BayesActiveSample1,BayesActiveSample2}) sections of audio data, a section much shorter than $5$ seconds would make the task of tuning into each new example very difficult due to the absence of a reference signal. 

%The trade-off with this approach is that the event often occurs on timescales different to that of the supplied weak labels, creating erroneously labelled data when viewed on a feature window level. This label error is precisely what we aim to correct. Furthermore, we assume that within the $5$ second chunks, any section which is labelled as an event ($1$) contains a signal, and any that is labelled as a non-event ($0$) does not.

% Talk about the noise and event signal

%For all inputs the features are normalised to prevent inaccurate results due to sensitivity to scaling for various algorithms. The log-mel feature space is further compressed by choosing $7$ features using the two-sample KS test as detailed in Section \ref{KS}. This stems from the requirement of a balance between the ease of separability of the classes, which is obtained in lower dimensions, whilst maintaining sufficient complexity in our feature space.

\subsection{Experimental Design}
We evaluate our inner model against the baseline classifiers with two experiments and finally test the overall performance of the framework utilising the outputs of the various inner classifiers.  We make the assumption that the accuracy of the weak labels is $100\%$. %Our initial investigation has shown that relaxing this assumption induces a minor error rate. However, this error rate is reasonably consistent between methods, not affecting conclusions drawn from the main thread of work. We therefore leave the effect of label error rate as further work
Therefore, all the classifiers predict over the coarse class $1$ labelled data only. Additionally, the priors we use in Eq. \ref{eq:Bayes} for our Gaussian KDE model are set such that $p(C_{0})=p(C_{1})=0.5$. This is to reflect that, since the audio sample is weakly labelled $1$, each data point is equally likely to be in fine class $0$ or $1$. 

Generative models, such as the Gaussian KDE obtain a performance boost from the additional information provided by the coarse class $0$ data as this allows it to better model the class $0$ distribution. Conversely, discriminative models such as the SVM, RF and MLP take a hit in performance because the decision boundary that they create over-fits to the class $0$ data points due to the increased class imbalance. We therefore train the LDA, GNB, SVM, RF and MLP on the finely-labelled data only, whereas the Gaussian KDE is trained on both the finely-labelled data and the coarse class $0$ data.

\section{Classification Performance}
\label{Classification Performance}
\begin{figure*}
\centering
  \includegraphics[width=1.0\textwidth]{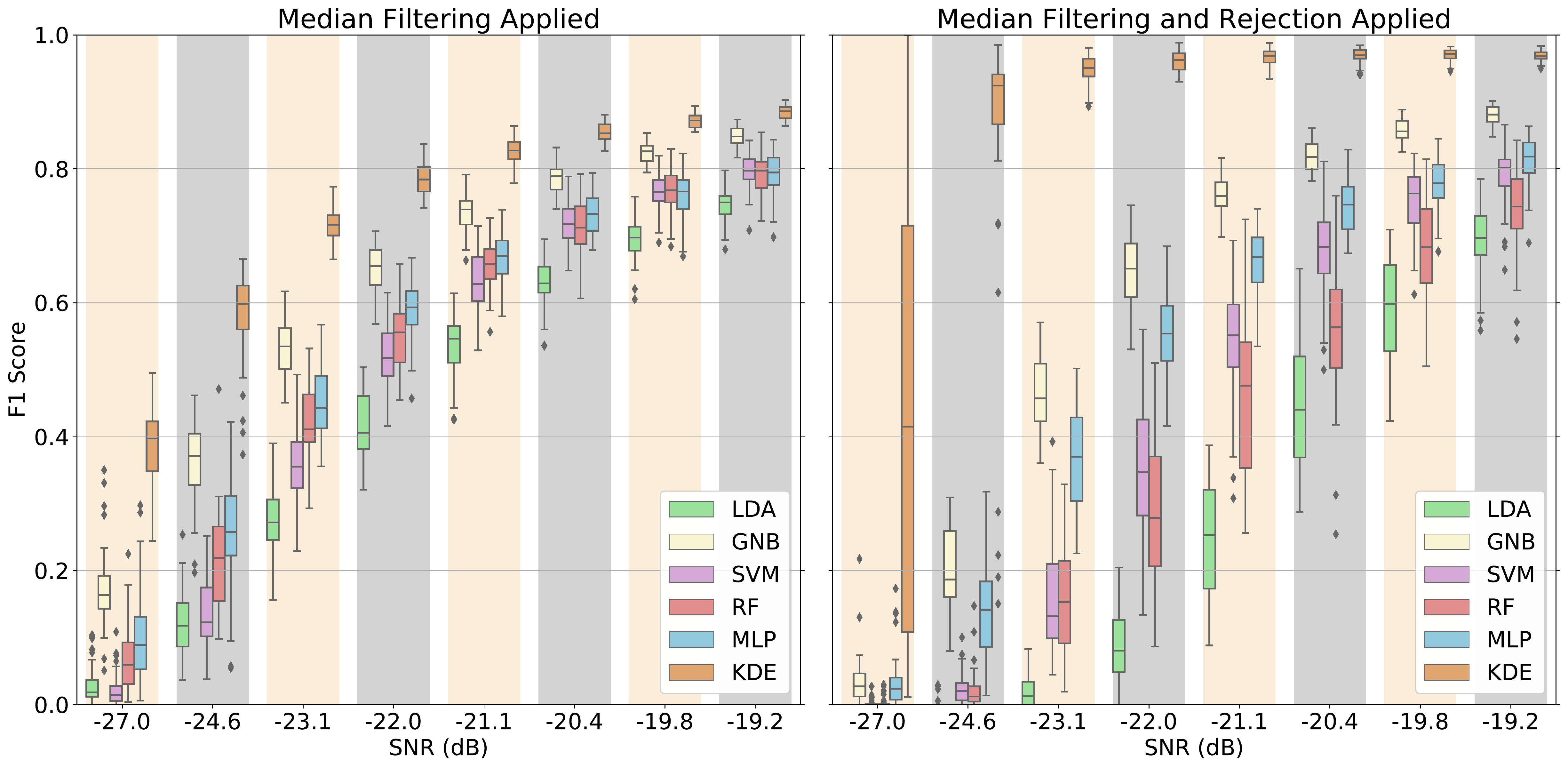}  \caption{From left to right: Boxplots showing results of first two experiments,  grouped by SNR. LDA, SVM, RF and MLP are trained on finely-labelled data only whilst the Gaussian KDE is trained on the finely-labelled data and coarse class $0$ data.}
\label{fig:PostProcresults}
\end{figure*}

% \begin{enumerate}[leftmargin=*]
    % \item
    For each SNR we run $40$ iterations, varying the time location of the injected signals, as well as the random seed of the algorithms. After applying median filtering, with a filter window of $500$ ms, we see the results in Fig. \ref{fig:PostProcresults}. 
    The F$1$-score gradually increases as expected from the threshold of detection to more audible SNRs. The decay of performance at the lower SNRs can be partially accounted for by the two-sample KS test used for feature selection failing to choose features of interest due to the increased noise floor. We observe a significant benefit to using the Gaussian KDE, which when combined with temporal averaging helps recover the dynamic nature of the signal (namely that there is correlation between neighbouring values in the time-series).     
    
    Fig. \ref{fig:PostProcresults} shows that the Gaussian KDE predicts better calibrated probabilities than the other baseline classifiers. This is shown by applying rejection \cite{NueralNet, rejection} in addition to the median filtering. The
    rejection window for the output probabilities is $0.1<p(C_{1}| \bm{x})<0.9$. The Gaussian KDE improves significantly in performance, especially at the lower SNRs; however it should be noted that the F$1$-score is evaluated on the remaining data after rejection. %The proportion of data rejected can be seen in Fig. \ref{fig:Ratio} in Appendix \ref{ExpAppendix}. 
    The Gaussian KDE rejects a large proportion of the data at lower SNRs, showing that the probabilities are at either extremes only when the model is confident in its predictions.

 The final experiment tests the overall framework with input to a CNN from pseudo-finely labelled data with median filtering and rejection applied. Table \ref{table:CNNResults} shows that using the framework in conjunction with any of the inner classifiers outputs outperforms a regular CNN trained on the coarse data. Furthermore, training the CNN on the output of the Gaussian KDE significantly improves detection of events by $22.1\%$ over the best baseline system, the CNN(GNB). We also show that using the strongest inner classifier (KDE) alone results in vastly lower precision and recall scores to the bootstrapping approach advocated here, which sees an improvement of $0.22$ to the F$1$-score gained by incorporating the CNN into the pipeline with the KDE. 
% \end{enumerate}

\setlength{\tabcolsep}{4.5pt}
\begin{table}[!h]
\begin{center}
\caption{CNN outer classifier: Metrics reported as means $\pm$ one standard deviation at an SNR of $-19.8$ dB for $40$ iterations}
\resizebox{0.48\textwidth}{!}{%

\begin{tabular}{cccc}
\hline
Classifier & F$1$-score & Precision & Recall \\
\hline
\textbf{CNN(KDE)} & \textbf{0.729 $\pm$ 0.034} & \textbf{0.719 $\pm$ 0.029} & \textbf{0.744 $\pm$ 0.031}\\
CNN(MLP)          & $0.435 \pm 0.022$ & $0.667 \pm 0.026$ & $0.322 \pm 0.029$\\
CNN(RF)           & $0.320 \pm 0.031$ & $0.419 \pm 0.035$ & $0.259 \pm 0.034$\\
CNN(SVM)          & $0.338 \pm 0.024$ & $0.484 \pm 0.024$ & $0.259 \pm 0.022$\\
CNN(GNB)          & $0.597 \pm 0.023$ & $0.654 \pm 0.028$ & $0.549 \pm 0.023$\\
CNN(LDA)          & $0.307 \pm 0.027$ & $0.571 \pm 0.026$ & $0.210 \pm 0.023$\\
CNN(Coarse)       & $0.174 \pm 0.036$ & $0.095 \pm 0.031$ & $0.923 \pm 0.039$\\
\hline
KDE               & $0.506 \pm 0.021$ & $0.518 \pm 0.021$ & $0.502 \pm 0.024$\\
\hline
\label{table:CNNResults}
\end{tabular}}
\end{center}
\vspace{-9mm}
\end{table}
\FloatBarrier
\section{Conclusions \& Further Work}
\label{Conclusion}

\subsection{Conclusions} 
This paper proposes a novel framework utilising a Gaussian KDE for super-resolving weakly-labelled data to be fed into a CNN to predict over unlabelled data. Our framework is evaluated on synthetic data and achieves an improvement of $22.1\%$ in F$1$-score over the best baseline system. We thus highlight the value label super-resolution provides in domains with only small quantities of finely-labelled data, a problem in the literature that is only sparsely addressed to date.

\subsection{Further Work}

To leverage the probabilistic labels outputted by the inner classifier, a suitable candidate for the outer classifier is a loss-calibrated Bayesian neural network (LC-BNN). This combines the benefits of deep learning with principled Bayesian probability theory \cite{LC-BNN}.  

Due to computational limitations, optimisation of the hyper-parameters was infeasible. Future work plans to use Bayesian Optimisation \cite{BaysOpt} for this tuning.

Finally, following the promising results of this paper, the next step is application to real datasets.

\FloatBarrier


\begin{thebibliography}{99}
\bibliographystyle{plainnat}

% \bibliography{example_paper}
% \bibliographystyle{icml2019}
% Introduction
\bibitem[Bishop(1995)]{NueralNet}
Bishop, C.M., (1995). Neural networks for pattern recognition. Oxford University Press.


\bibitem[Bottou et al.(2010)]{SGD}
Bottou, L., (2010). Large-scale machine learning with stochastic gradient descent. In \textit{Proceedings of COMPSTAT'2010}, pp. 177-186.

\bibitem[Cartwright et al.(2019)]{CrowdSourcing1}
Cartwright, M., Dove, G., Méndez, A.E.M. and Bello, J.P., (2019). Crowdsourcing multi-label audio annotation tasks with citizen scientists. In \textit{Proceedings of the 2019 CHI Conference on Human Factors in Computing Systems}, pp. 4-9.


\bibitem[Cobb et al.(2018)]{LC-BNN}
Cobb, A.D., Roberts, S.J. and Gal, Y., (2018). Loss-calibrated approximate inference in Bayesian neural networks. In \textit{arXiv preprint arXiv:1805.03901}


\bibitem[Deng et al.(2009)]{CrowdSourcing2}
Deng, J., Dong, W., Socher, R., Li, L.J., Li, K. and Fei-Fei, L., (2009). Imagenet: A large-scale hierarchical image database. In \textit{2009 IEEE conference on computer vision and pattern recognition}, pp. 248-255.


\bibitem[Hanczar and Dougherty(2008)]{rejection}
Hanczar, B. and Dougherty, E.R., (2008). Classification with reject option in gene expression data. In \textit{Bioinformatics}, 24(17):1889-1895.

\bibitem[Hayashi et al.(2017)]{LogMel}
Hayashi, T., Watanabe, S., Toda, T., Hori, T., Le Roux, J. and Takeda, K., (2017). BLSTM-HMM hybrid system combined with sound activity detection network for polyphonic sound event detection. In \textit{2017 IEEE International Conference on Acoustics, Speech and Signal Processing (ICASSP)}, pp. 766-770.


\bibitem[Houlsby et al.(2011)]{BayesActiveSample1}
Houlsby, N., Huszár, F., Ghahramani, Z. and Lengyel, M., (2011). Bayesian active learning for classification and preference learning. In \textit{arXiv preprint arXiv:1112.5745}.

\bibitem[Ivanov et al.(2012)]{KS1}
Ivanov, A. and Riccardi, G., (2012). Kolmogorov-Smirnov test for feature selection in emotion recognition from speech. In \textit{2012 IEEE international conference on acoustics, speech and signal processing (ICASSP)}, pp. 5125-5128

\bibitem[Kiskin et al.(2018)]{Wavelet}
Kiskin, I., Zilli, D., Li, Y., Sinka, M., Willis, K. and Roberts, S., (2018). Bioacoustic detection with wavelet-conditioned convolutional neural networks. In \textit{Neural Computing and Applications}, pp.1-13

\bibitem[Kong et al.(2019)]{LogMel4}
Kong, Q., Xu, Y., Sobieraj, I., Wang, W. and Plumbley, M.D., (2019). Sound Event Detection and Time–Frequency Segmentation from Weakly Labelled Data. In \textit{IEEE/ACM Transactions on Audio, Speech and Language Processing (TASLP)}, 27(4):777-787.

\bibitem[Krizhevsky et al.(2012)]{DNNSmallData1}
Krizhevsky, A., Sutskever, I. and Hinton, G.E., (2012). Imagenet classification with deep convolutional neural networks. In \textit{Advances in neural information processing systems}, pp. 1097-1105.

\bibitem[LeCun et al.(2015)]{NN}
LeCun, Y., Bengio, Y. and Hinton, G., (2015). Deep learning. In \textit{Nature}, 521(7553):436.


\bibitem[Li et al.(2017)]{HumBug}
Li, Y., Zilli, D., Chan, H., Kiskin, I., Sinka, M., Roberts, S. and Willis, K., (2017). Mosquito detection with low-cost smartphones: data acquisition for malaria research. In \textit{arXiv preprint arXiv:1711.06346}.


\bibitem[Lin et al.(2014)]{CrowdSourcing3}
Lin, T.Y., Maire, M., Belongie, S., Hays, J., Perona, P., Ramanan, D., Dollár, P. and Zitnick, C.L., (2014). Microsoft coco: Common objects in context. In \textit{European conference on computer vision}, pp. 740-755.


\bibitem[Naghshvar et al.(2012)]{BayesActiveSample2}
Naghshvar, M., Javidi, T. and Chaudhuri, K., (2012). Noisy bayesian active learning. In \textit{2012 50th Annual Allerton Conference on Communication, Control, and Computing (Allerton)}, pp. 1626-1633

\bibitem[Parzen et al.(1962)]{KDE2}
Parzen, E., (1962). On estimation of a probability density function and mode. In \textit{The annals of mathematical statistics}, 33(3):1065-1076.


\bibitem[Rolnick et al.(2017)]{DNNSmallData2}
Rolnick, D., Veit, A., Belongie, S. and Shavit, N., (2017). Deep learning is robust to massive label noise. In \textit{arXiv preprint arXiv:1705.10694}.


% % Methodlogy
% % Gaussian KDE
\bibitem[Scott(2015)]{KDE1}
Scott, D.W., (2015). Multivariate density estimation: theory, practice, and visualization. John Wiley \& Sons.


% % Conclusion
\bibitem[Snoek et al.(2012)]{BaysOpt}
Snoek, J., Larochelle, H. and Adams, R.P., (2012). Practical {B}ayesian optimization of machine learning algorithms. In \textit{Advances in neural information processing systems}, pp. 2951-2959

\bibitem[Srivastava et al.(2014)]{Dropout}
Srivastava, N., Hinton, G., Krizhevsky, A., Sutskever, I. and Salakhutdinov, R., (2014). Dropout: a simple way to prevent neural networks from overfitting. In \textit{The Journal of Machine Learning Research}, 15(1):1929-1958.





\bibitem[Zooniverse(2019)]{Zooniverse}
Zooniverse Humbug page (2019), \url{https://www.zooniverse.org/projects/yli/humbug}. Accessed: 2019-04-29





\end{thebibliography}
\end{document}